\documentclass[10pt,twocolumn,letterpaper]{article}

\usepackage{cvpr}              
\definecolor{cvprblue}{rgb}{0.21,0.49,0.74}
\usepackage[pagebackref=False,breaklinks,colorlinks,allcolors=cvprblue]{hyperref}
\usepackage{multirow,xcolor,colortbl,tabularx,array}
\usepackage{float}
\usepackage{caption}
\usepackage{graphicx}%
\usepackage{amsmath,amssymb,amsfonts}%
\usepackage{amsthm}%
\usepackage{mathrsfs}%
\usepackage[title]{appendix}%
\usepackage{textcomp}%
\usepackage{manyfoot}%
\usepackage{booktabs}%
\usepackage{algorithm}%
\usepackage[noend]{algpseudocode}%
\usepackage{listings}%
\usepackage{multicol}
\usepackage{makecell}
\usepackage{bm}
\usepackage{CJKutf8}
\usepackage{hyperref}

\graphicspath{{figs/}}
\newcolumntype{Y}{>{\centering\arraybackslash}X}

\definecolor{Group}{RGB}{245,245,245}

\def\paperID{44768} 
\def\confName{CVPR}
\def\confYear{2026}

\title{ForestPrune: High-ratio Visual Token Compression for Video Multimodal Large Language Models via Spatial-Temporal Forest Modeling}

\author{Shaobo Ju$^{1}$\footnotemark[1], Baiyang Song$^{1}$\thanks{Equal contribution.}, Tao Chen$^{1}$, Jiapeng Zhang$^{1}$, Qiong Wu$^{1}$,\\ Chao Chang$^{2}$, HuaiXi Wang$^{2}$, Yiyi Zhou$^{1}$\thanks{Corresponding Author.}, Rongrong Ji$^{1}$\\
$^{1}$ Key Laboratory of Multimedia Trusted Perception and Efficient Computing,\\
Ministry of Education of China, Xiamen University, 361005, P.R. China.\\
$^{2}$ National University of Defense Technology.\\
{\tt\small \{jushaobo,songbaiyang,chentao,zhangjiapeng,qiong\}@stu.xmu.edu.cn, \{zhouyiyi, rrji\}@xmu.edu.cn,}\\
{\tt\small \{changchao,wanghuaixi\}@nudt.edu.cn}
}

\begin{document}
\maketitle
\begin{abstract}
Due to the great saving of computation and memory overhead, token compression has become a research hot-spot for MLLMs and achieved remarkable progress in image-language tasks. However, for the video, existing methods still fall short of high-ratio token compression. We attribute this shortcoming to the insufficient modeling of temporal and continual video content, and propose a novel and training-free token pruning method for video MLLMs, termed \emph{\textbf{ForestPrune}}, which achieves effective and high-ratio pruning via Spatial-temporal Forest Modeling. In practice, ForestPrune construct token forests across video frames based on the semantic, spatial and temporal constraints, making an overall comprehension of videos. Afterwards, ForestPrune evaluates the importance of token trees and nodes based on tree depth and node roles, thereby obtaining a globally optimal pruning decision. 
To validate ForestPrune, we apply it to two representative video MLLMs, namely \emph{LLaVA-Video} and \emph{LLaVA-OneVision}, and conduct extensive experiments on a bunch of video benchmarks. 
The experimental results not only show the great effectiveness for video MLLMs, \emph{e.g.}, retaining 95.8\% average accuracy while reducing 90\% tokens for LLaVA-OneVision, but also show its superior performance and efficiency than the compared token compression methods, \emph{e.g.}, +10.1\% accuracy on MLVU and -81.4\% pruning time than FrameFusion on LLaVA-Video. 
\textbf{Our} \textbf{code} is given in the \href{https://github.com/luminousllsa/ForestPrune}{\textbf{ForestPrune}}.
\end{abstract}    
\vspace{-0.4cm}
\begin{figure}[t]
    \centering
    \includegraphics[width=\linewidth]{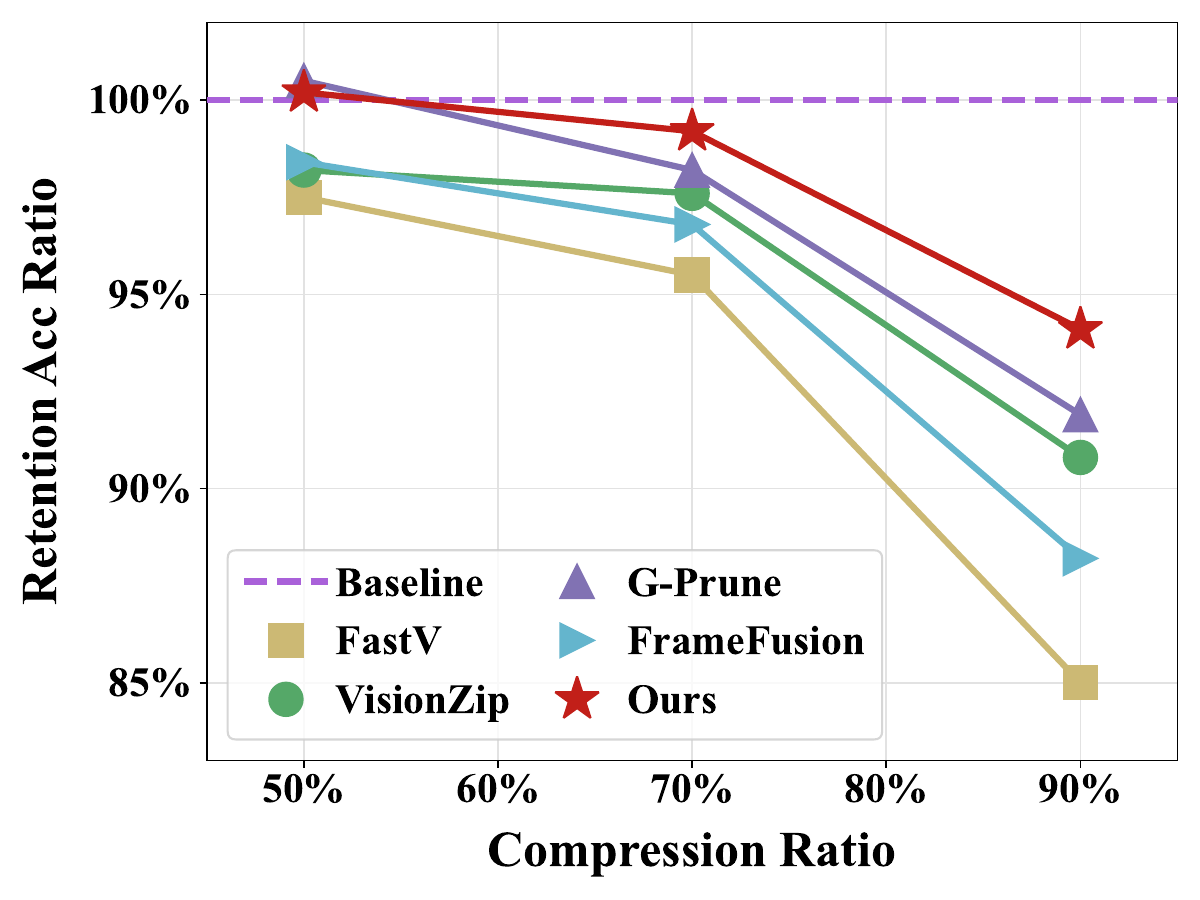} 
    \vspace{-0.5cm}
    \setlength{\abovecaptionskip}{0.2cm}
    \caption{Comparison between the proposed \emph{ForestPrune} (Ours) and the existing compression methods ~\cite{jiang2025kind, fastv, yang2025visionzip, fu2024framefusion} on VideoMME. The base model used is LLaVA-Video-7B. As pruning ratio increases, existing methods will encounter obvious performance drops, while ForestPrune is still robust.}
    \label{fig:motivation}
    \vspace{-0.5cm}
\end{figure}
\section{Introduction}
\label{sec:intro}
\vspace{-0.2cm}

Recent years have witnessed the rapid development of \emph{multimodal large language models} (MLLMs)~\cite{blip, blip-2, instructblip, gpt4, clip, internvl-1.5, llava1.5, llavahr, llava, llamavid, llavanext} and their great breakthroughs made in a variety of vision-language tasks~\cite{in, luo1, luo2, luo3, abs-2407-15047, RadenovicDKMVPW23}. 
Despite the great success, high latency and excessive computation overhead remain open problems that plague the application of MLLMs. 
In pursuit of stronger visual understanding, recent MLLMs~\cite{llava1.5, llavahr, llava} typically use a large number of visual tokens to represent a single image, leading to a quadratic increase in computation and obvious visual redundancy~\cite{weng2024longvlm, chai2024auroracap}. 
These cases become much more prominent for MLLMs when handling the video-based tasks, where dozens to hundreds of image frames are used as the model input~\cite{llava-next-video, videollava, video-chatgpt, videochat, Llava-onevision, zhang2024video, Qwen2-VL, Qwen2.5-vl, Internvl2.5, Internvl3.5}.


In this case, numerous efforts~\cite{fastv, tome, yang2025visionzip, yang2025topv, jiang2025kind, ye2024fit, wu2025routing, wu2024accelerating, kong2026tokenreductionefficiencygenerative} have recently been devoted to the exploration of visual token compression for MLLMs, so as to reduce the expenditure of both computation and memory overhead while retaining high performance. 
These endeavors often resort to the principles of token pruning~\cite{fastv, yang2025topv, ye2024fit} or merging~\cite{tome, yang2025visionzip, shang2025llava}, and design approaches based on the metrics of visual saliency~\cite{yang2025topv, xing2024pyramiddrop} and diversity~\cite{jiang2025kind, zou2025don} or cross-modal relevance~\cite{man2025adacm, zhang2024sparsevlm}. 
In this case, existing compression methods can reduce the visual redundancy via retaining the most important tokens, which have achieved remarkable success for image-based MLLMs and tasks~\cite{llava1.5, llavanext, goyal2017making, hudson2019gqa, liu2024mmbench}.

\begin{figure}[t]
    \centering
    \includegraphics[width=\linewidth]{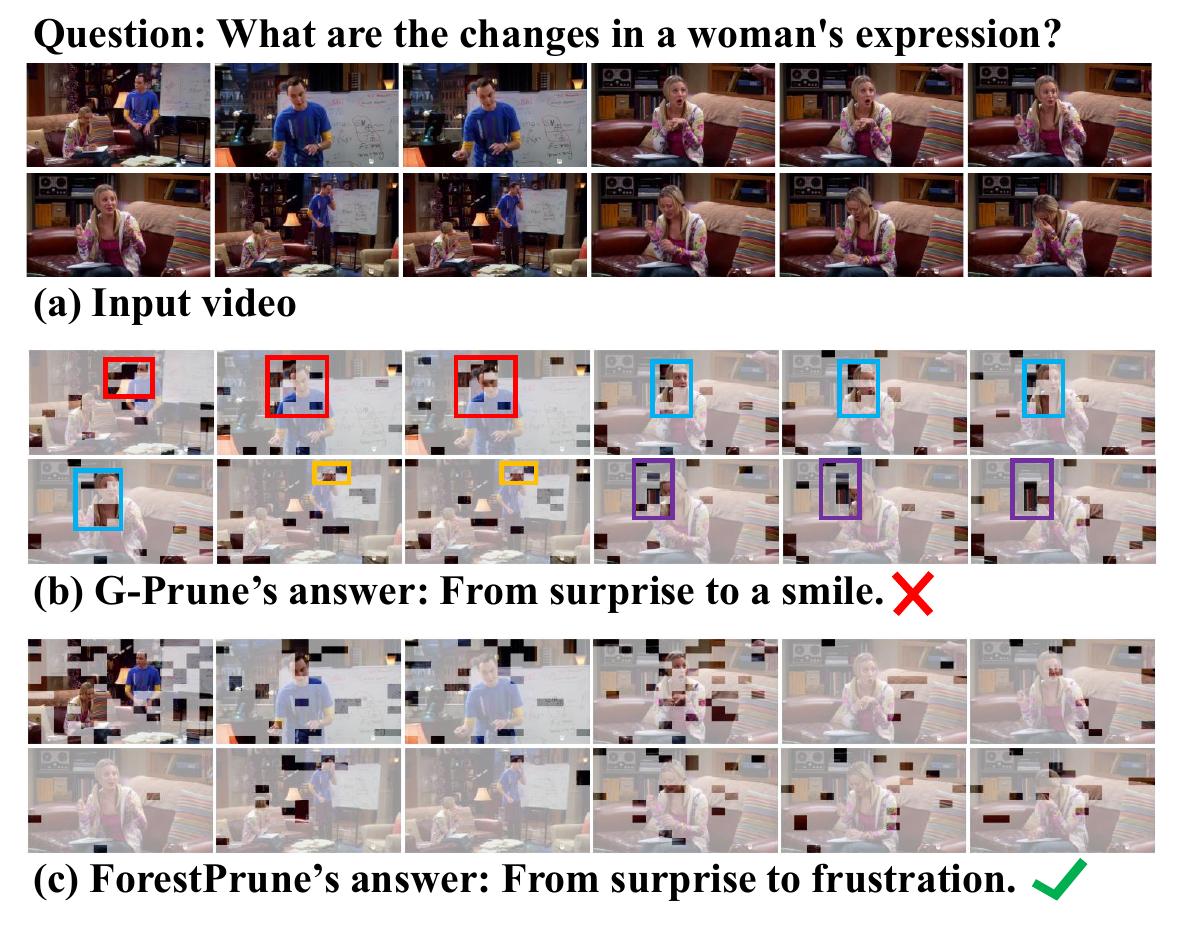}
    \vspace{-0.5cm}
    \setlength{\abovecaptionskip}{0.2cm}
    \caption{Visualization of token pruning by G-Prune ~\cite{jiang2025kind} and our ForestPrune. The image-centric G-Prune can well keep the important tokens for each frame, but also leads to obvious redundancy across frames. In contrast, our ForestPrune can obtain a globally optimal pruning via spatial-temporal forest modeling.}
    \label{fig:visual_intro}
    \vspace{-0.5cm}
\end{figure}

However, for the video tasks, existing compression methods still have ample room to improve, especially considering the high-ratio compression settings.
As shown in Fig \ref{fig:motivation}, existing compression methods, such as G-Prune~\cite{jiang2025kind} and VisionZip~\cite{yang2025visionzip}, can still maintain high performance while compressing about half of the visual tokens on the video benchmarks like Video-MME~\cite{Video-mme}. 
However, as the compression ratio increases, their performance gap becomes much more obvious, distinct from their effects for image-based tasks~\cite{goyal2017making, hudson2019gqa, liu2024mmbench}. 
This case can be attributed to the insufficient modeling of video content. 
Also shown in Fig \ref{fig:visual_intro}, under the high compression ratio, the retained visual tokens of G-Prune in adjacent frames are prone to similarity and redundancy, and this case suggests that existing compression methods mainly focus on image-wise importance, while lacking a sufficient evaluation of global visual redundancy across video frames.

In this paper, we investigate this problem from the perspective of \emph{spatial-temporal forest modeling}, and propose an innovative and training-free method termed \emph{ForestPrune} for video MLLMs. 
In principle, ForestPrune aims to model the importance of visual tokens across frames via building semantic trees based on both their spatial and temporal priors. 
In this case, we can evaluate the global importance of visual tokens based on the depth of trees and the role of nodes, \emph{e.g.}, the root or trunk nodes of a deeper tree are more important. 
In practice, ForestPrune first obtains the representative tokens of each video frame and then considers them as the potential tree nodes. 
Afterwards, ForestPrune constructs a spatial-temporal forest by gating tokens with semantic similarity, spatial alignment and temporal order. 
Given the constructed token forests, we perform the budgeted selection of tokens via prioritizing the pruning of leaf and tail nodes. 
In this case, ForestPrune can obtain a global optimal pruning decision across video frames. 

To validate ForestPrune, we apply it to two representative MLLMs, namely LLaVA-Video~\cite{zhang2024video} and LLaVA-OneVision~\cite{Llava-onevision}, and conduct extensive experiments on five highly competitive video benchmarks~\cite{NExT-QA, Video-mme, MLVU, LongVideoBench, Mvbench}. 
The experimental results not only show the outstanding compression results of ForestPrune for two MLLMs, \emph{e.g.}, retaining 95.8\% average accuracy while pruning 90\% tokens of LLaVA-OneVision, but also show its superiority than the compared methods, \emph{e.g.}, +10.1\% accuracy than FrameFusion~\cite{fu2024framefusion} on MLVU while reducing 81.4\% more computation on LLaVA-Video. 
Moreover, using ForestPrune to scale up input frames can also help LLaVA-Video approach SOTA performance on some metrics, \emph{e.g.}, 72.5 on MLVU. 
These results well validate the effectiveness of our ForestPrune for video-MLLMs in terms of high-ratio visual compression, and also confirm our intuition about spatial-temporal modeling for video redundancy evaluation. 
\\Overall, our contributions are three-fold:

\begin{itemize}
    \item We reveal the critical ingredient for effective video token compression, \emph{i.e.}, the spatial and temporal modeling for continuous video content. 
    \item We propose a novel and training-free approach for video MLLMs, termed \emph{ForestPrune}, which adopts spatial-temporal forest modeling to evaluate the global importance of visual tokens across video frames. 
    \item The proposed ForestPrune helps two video-MLLMs well retain the performance while reducing a large number of redundant tokens. Moreover, it also shows obvious merits than the compared compression methods in terms of both performance and efficiency.
\end{itemize}
\section{Related Work}
\label{sec:related work}
\vspace{-0.1cm}
\subsection{Video LLMs}
\vspace{-0.2cm}
Recent multimodal LLMs have been extended to videos~\cite{gpt4, GPT-4O, llama, llamavid, video-llama, videollama2, llava, llava-next-video, Llava-onevision, zhang2024video, Qwen2-VL, Qwen2.5-vl, Internvl2.5, Internvl3.5, nvila}. LLaVA-OneVision~\cite{Llava-onevision} unifies image and video in a single framework, transferring visual knowledge across scenarios with high-resolution inputs. LLaVA-Video~\cite{zhang2024video} scales to large synthetic video-instruction data, yielding strong results across diverse video QA/reasoning benchmarks. Beyond LLaVA family models, Qwen2-VL~\cite{Qwen2-VL} emphasizes high-resolution encoders and long-context handling, Qwen2.5-VL~\cite{Qwen2.5-vl} upgrades Qwen2-VL with stronger recognition and long-video comprehension under a unified image–video paradigm. NVILA~\cite{nvila} targets efficiency by scaling spatial and temporal resolution while compressing visual tokens for a better accuracy–latency trade-off. Despite these advances, processing multiple frames leads to a surge in visual tokens and heavy compute costs, motivating efficient token compression for video LLMs. InternVL 3.5~\cite{Internvl3.5} adopts a Visual Resolution Router and Decoupled Vision-Language Deployment to significantly improve the performance of open-source video MLLMs.

\subsection{Visual Token Compression Methods}
\vspace{-0.2cm}
Visual token compression methods were initially developed for image tasks and later extended to video domains~\cite{fastv, yang2025topv, tome, yang2025visionzip, ye2025atp, zhao2025accelerating, yang2025pvc, alvar2025divprune, dhouib2025pact, tao2025dycoke, zhou2019plenty, wu2026not}. FastV~\cite{fastv} drops low-contribution visual tokens using early-layer attention scores to reduce computation while keeping accuracy. VisionZip~\cite{yang2025visionzip} first selects a small set of anchor tokens via saliency and then semantically merges the remaining tokens into compact composites. FitPrune~\cite{ye2024fit} selects retained tokens by minimizing the attention distribution difference before and after pruning. {G-Prune}~\cite{jiang2025kind} builds a similarity graph and preserves a representative set of tokens by propagating importance to avoid redundant picks. HoloV~\cite{zou2025don} allocates the pruning budget across spatial regions to prevent repeatedly keeping highly similar “hotspot” tokens. For video visual token compression methods, PruneVid~\cite{huang2025prunevid} leverages LLMs’ reasoning capabilities to selectively prune visual features which are relevant to question. FrameFusion~\cite{fu2024framefusion} aligns adjacent frames to merge duplicated regions early and prunes by importance in deeper layers to balance speed and accuracy. STTM~\cite{sttm} performs multi-granularity merging within and across frames to explicitly model spatial–temporal redundancy. Holitom~\cite{shao2025holitom} employs outer-LLM pruning through global redundancy-aware temporal segmentation, followed by spatial-temporal merging to reduce visual tokens. 
Overall, existing methods are more effective for single images but also easy to keep redundancy across frames. 
\begin{figure*}[t]
    \centering
    \includegraphics[width=\textwidth]{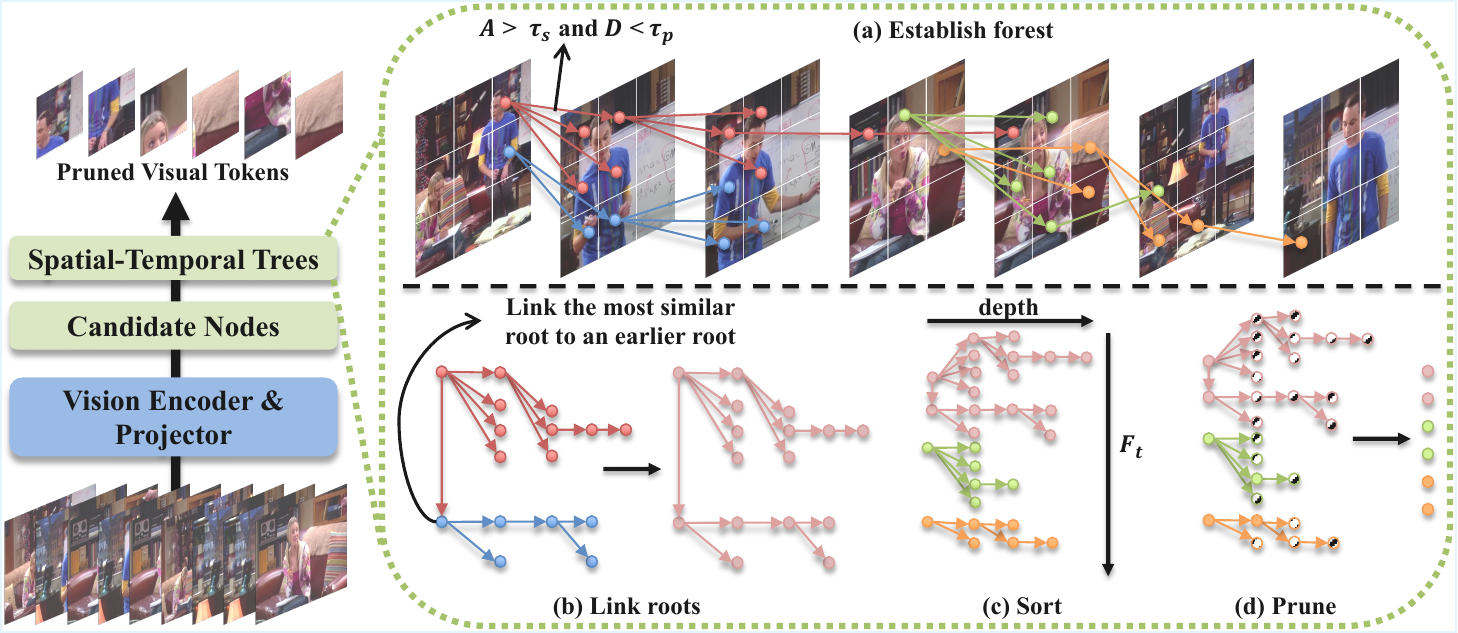} 
    \vspace{-0.5cm}
    \setlength{\abovecaptionskip}{0.2cm}
    \caption{Illustration of the proposed ForestPrune. 
    Input video frames are first encoded by the visual encoder, based on which ForestPrune will select a set of tokens of each frame as the candidate nodes. Afterwards, ForestPrune constructs the token trees based on the semantic similarity $\tau_s$, spatial distance $\tau_p$ and the frame temporal orders, thereby forming the \emph{spatial-temporal forest} (a). When obtaining excessive trees (root nodes), we will merge them before pruning (b). Then, we sort the trees in a descending order of depth (c) and then progressively prune the leaf and tail nodes until meeting the compression budget (d). Via the spatial-temporal modeling, ForestPrune can well estimate the frame-wise redundancy and obtain a globally optimal pruning decision.}
    \label{fig:Method}
    \vspace{-0.5cm}
\end{figure*}
\section{Method}
\label{sec:method}
\vspace{-0.1cm}
\subsection{Overview}
\label{overview}
\vspace{-0.2cm}
In this paper, we propose an innovative and training-free method termed \emph{ForestPrune} for video MLLMs, of which framework is illustrated in Fig \ref{fig:Method}. 
ForestPrune aims to model the visual tokens across frames via building semantic trees based on both their spatial and temporal priors. 
In this case, we can evaluate the global importance of visual tokens based on the tree depth and node roles. 

Concretely, given a video $V$ sampled for $T$ frames, each frame is processed by the image encoder, and we obtain the frame features $\mathbf{F}_{t_i}\in \mathbb{R}^{N\times d}$, where $N$ denotes the number of tokens and $d$ is the feature dimension. Thus, the whole video features can be denoted by $\mathbf{F}_V\in\mathbb{R}^{(T\times N)\times d}$. 

In terms of visual compression, the optimization aims to reduce the number of visual tokens according to a predefined compression budget $b$:
\begin{equation}
\min \mathcal{D}\big(G(\mathbf{F}_V'),\ \ G(\mathbf{F}_V)\big) \ s.t. \ b.
\end{equation}
Here, $G(\cdot)$ denotes the MLLM and $\mathcal{D(\cdot)}$ denotes the difference between the outputs of MLLM before and after compression.
$\mathbf{F}_V' \in \mathbb{R}^{K\times d}$ is the compressed video features, which can be achieved via token pruning~\cite{jiang2025kind} or merging ~\cite{yang2025visionzip}. And the budget $b$ is calculated by $K/(T\times N)$.

Existing compression methods~\cite{jiang2025kind, yang2025visionzip} for image MLLMs often focus on image-wise token compression, \emph{i.e.}, $\mathbf{F}_{t_i} \in \mathbb{R}^{k\times d}$, where $k$ is the number of retained tokens per frame, yielding a multi-frame compression budget $b=T\times k/T\times N$. 
However, this paradigm is prone to losing the overall consideration of all visual tokens, and also lacks of temporal and continual modeling of videos.

In this case, ForestPrune models the token compression via \emph{spatial-temporal forests}. Concretely, for the better construction of node forests (trees), we first select the representative tokens of each frame as the nodes, of which process can be conducted via existing token pruning methods~\cite{yang2025visionzip, jiang2025kind} or random sampling. 

Thus, we can obtain the slimmed video features, denoted as $\mathbf{F}_{nod} \in \mathbb{R}^{(T\times N' ) \times d}$, where $N'< N$. 
Based on $\mathbf{F}_{nod}$, we then construct the semantic trees based on the semantic distance, spatial constraint  and temporal order, of which process will be detailed later. Then, we can obtain a spatial-temporal forest consisting of token trees, denoted by $\mathbf{F}_{tree}^i \in \mathbb{R}^{k\times d}$, where $k$ denotes the number of tree nodes.  

Afterwards, we rank the importance of these trees based on their depths, and then select the token nodes based on their roles, \emph{i.e.}, the root or trunk nodes. 
In this case, we can obtain the compressed video features $\mathbf{F}_v'\in \mathbb{R}^{K\times d}$, where $K<<T\times N$.
The detailed procedure of spatial-temporal tree construction and token selection is given in the following selections.

\vspace{-0.1cm}
\subsection{Spatial-Temporal Forest Construction}
\vspace{-0.2cm}
\label{sttree}
In terms of token tree construction, we not only measure the semantics among nodes~\cite{tome}, but also consider the spatial and temporal priors. 
In this case, we also record the coordinate information of each token in their default frames, 
$S\in \mathbb{R}^{(T\times N')\times \text{2}}$, as well as their temporal timesteps, 
$\mathrm{T}\in \mathbb{R}^{(T\times N')\times \text{1}}$, for spatial-temporal forest modeling .

Concretely, given the representative node features of all frames $\mathbf{F}_{nod}$, we first compute their adjacency matrix of cosine similarity $A\in\mathbb{R}^{(T\times N')\times (T \times N')}$:
\begin{equation}
A=\mathbf{F}_{nod}\mathbf{F}_{nod}^\text{T}.
\end{equation}
Meanwhile, we also compute the spatial distances between nodes, forming the distance matrix $D\in\mathbb{R}^{(T\times N') \times ({T\times N'})}$ based on $S$. And its element is
\begin{equation}
d_{ij} = \lVert s_{i} - s_{j} \rVert_2 \ ,
\end{equation}
where $s_i \in \mathbb{R}^2$ represents the coordinates of the $i$-th node.

Based on $A$ and $D$, we can build a connection matrix $C \in \mathbb{R}^{(T\times N')\times (T\times N')}$ between nodes thresholded by the semantic and temporal constraints, denoted as $\tau_s$ and $\tau_p$, respectively. The element of $C$ is defined by
\begin{equation}
c_{ij}=
\begin{cases}
    1\quad \text{if }\ (a_{ij}\geq\tau_s)\land(d_{ij}\leq\tau_p),\\
    0\quad \text{otherwise}.
\end{cases}
\label{eq:4}
\end{equation}

We further update $C$ based on the temporal timesteps of nodes $T$:
\begin{equation}
c_{ij}=
\begin{cases}
    1\quad \text{if}\ (c_{ij}=1)\land (t_{i}<t_{j}), \\
    0\quad \text{otherwise}.
\end{cases}
\label{eq:5}
\end{equation}
Via Eq.\ref{eq:4}-\ref{eq:5}, we can obtain the potential connections of nodes under the semantic, spatial and temporal constraints. However, the building of spatial-temporal trees remains unsolved.
To this end, we first obtain the set of root nodes via computing the input degrees of $C$:
\begin{equation}
    \begin{aligned}
        &\mathcal{R}=\Big\{j\mid j\in\mathcal{V}\land (\sum_{i=1}^{k} c'_{ij}=0) \land (\sum_{j=1}^{k} c'_{ij} \neq 0)\Big\}, \\
    \end{aligned}
    \label{eq:6}
\end{equation}
where $\mathcal{R}$ recodes the indices of root nodes, $\mathcal{V}$ denotes the set of token nodes, and $|\mathcal{V}|=k=T\times N'$.

Afterwards, we collect the child nodes for each root one. In particular, although $C$ reflect the feasible connections between nodes, but they are still likely linked to different trees (root nodes). To avoid this case, we build another ranking matrix $P$ based on $A$ and $D$:
\begin{equation}
P=C\ \odot\ \big(A - \lambda\,D\big),
\end{equation}
where $A-\lambda D$ is used to consider both semantic and spatial distances between nodes, their values are larger than 0. $\lambda$ is a hyper-parameter that controls the trade-off.

$P$ can be used to identify the belonging of child nodes via ranking their scores to root ones. 
Then, we identify the nodes for each root $j$ in $\mathcal{R}$ based on $P$ and $C$, and obtain the tree nodes $\mathcal{V}_j$:
\begin{equation}
\mathcal{V}_j=\Big\{ i \mid \arg\max_{i\in\mathcal{V}} p_{ij} \land (\sum_{i=1}^{k} p_{ij} \neq 0) \Big\}.
\end{equation}

Given the nodes of a tree $\mathcal{V}_j$, we use a \emph{LinkNode} function to obtain its connections $E$ and the tree depth $d$:
\begin{equation}
LinkNode( r_j, \mathcal{V}_j, T, C)\rightarrow E, d. 
\end{equation}
Afterwards we can obtain  the tree as well as its connections $\mathcal{T}_j=\langle V_j,E_j \rangle$. The detailed process of LinkNode is given in Algorithm \ref{alg:spatial-temporal-tree}, which well considers the temporal information of all nodes.
\begin{algorithm}[t]
\caption{LinkNode$(\cdot)$}
\label{alg:spatial-temporal-tree}
\begin{algorithmic}[1]
\Require $r_j$, $\mathcal{V}_j$, $T$, $C$
\Ensure $E$, $d$
\For{$t=1$ \textbf{to} $T-1$}
    \State $\mathcal{V}'_t=\Big\{n_i^t \mid n_i \in \mathcal{V}_j \land T_i = t \Big\}$
    \If{$\mathcal{V}'_t=\varnothing$}
        \State \textbf{continue}
    \EndIf
    \For{\textbf{each} $n_i^t \in \mathcal{V}'_t$}
        \If{$n_i^t = r_j$}
            \State $d=0$; \textbf{continue}
        \Else
            \State $E = E + \langle n_i^t,n_j^{t'} \rangle \ s.t. \ c_{ij} = 1 ,\ 1\leq t'<t$
            \State $d = d + 1$
        \EndIf
    \EndFor
\EndFor
\State \Return $E$, $d$
\end{algorithmic}
\end{algorithm}
In particular, when the number of root nodes are much larger than that of the retained tokens, \emph{i.e.}, $|\mathcal{R}|>>K$, we will combine trees based on the root similarities, and then update the new trees.

\vspace{-0.1cm}
\subsection{Token Pruning}
\label{token selection}
\vspace{-0.2cm}

After the construction of spatial-temporal forest, \emph{i.e.}, Sec. \ref{sttree}, we then conduct the token pruning based on the depth of trees and the role of nodes.

Specifically, given the depths of all trees in the forest, $d=\Big\{d_1,d_2,...,d_{|\mathcal{R}|}\Big\}$, we first sort the trees based on their depth, and then remove the leaf nodes from the trees in descending order of depth. When only root and trunk nodes remain in all trees, but the pruning budget is still not met. We will progressively remove the nodes at the ends of trees. 

For instance, given a tree denoted as $\mathcal{T}_j=\langle V_j,E_j\rangle$ with a depth of $d_j$, the leaf nodes $V_j'$ can be obtained by
\begin{equation}
\mathcal{V}'_j=\Big\{i \mid i\in \mathcal{V}_j \land (\sum_{j=1}^kc_{ij}=0) \Big\}, 
\end{equation}
and we discard the leaf nodes by $\mathcal{V}_k = \mathcal{V}_j - \mathcal{V}'_j$.

And for the tail nodes of the pruned tree, we can remove them via 
\begin{equation}
\mathcal{V}_{tail}=\max_{j\in\mathcal{V}_j} T_j, \  \mathcal{V}_k=\mathcal{V}_j - \mathcal{V}_{tail}.
\end{equation}

This process will be progressively conducted until reaching the pruning budget. Besides, under high-ratio pruning setting, if only the root nodes are kept, we will select the ones with default earlier timesteps as the remaining nodes for the budget. Finally, we obtain the pruned node set $\mathcal{V}_K \subset \mathcal{V}$ and the pruned feature $\mathbf{F}_{v}' \in \mathbb{R}^{K\times d}$.
\section{Experiments}
\label{sec:experiments}
\begin{table*}[t]
\centering
\scriptsize
\setlength{\abovecaptionskip}{0.1cm}
\setlength{\belowcaptionskip}{0.2cm}
\caption{
Comparison between ForestPrune and existing compression methods on LLaVA-Video-7B and LLaVA-OV 7B. \textbf{Compression Ratio} denotes the target reduction of visual tokens. \textbf{Retain} reflects the degree to which MLLM retains performance.
}
\label{tab:main_results}
\resizebox{\linewidth}{!}{
\begin{tabularx}{\linewidth}{@{} c c *{11}{Y} @{}}
\toprule
\multirow[c]{2}{*}{\makecell{\textbf{Compression}\\\textbf{Ratio}}} &
\multirow[c]{2}{*}{\textbf{Method}} &
\multicolumn{2}{c}{\textbf{NExT-QA}} &
\multicolumn{2}{c}{\textbf{VideoMME}} &
\multicolumn{2}{c}{\textbf{MLVU}} &
\multicolumn{2}{c}{\textbf{LongVideoBench}} &
\multicolumn{2}{c}{\textbf{MVBench}} &
\multicolumn{1}{c}{\textbf{Avg}} \\
\cmidrule(lr){3-4}\cmidrule(lr){5-6}\cmidrule(lr){7-8}\cmidrule(lr){9-10}\cmidrule(lr){11-12}\cmidrule(lr){13-13}
& & \textbf{Acc}$\uparrow$ & \textbf{Retain}$\uparrow$
& \textbf{Acc}$\uparrow$ & \textbf{Retain}$\uparrow$
& \textbf{Acc}$\uparrow$ & \textbf{Retain}$\uparrow$
& \textbf{Acc}$\uparrow$ & \textbf{Retain}$\uparrow$
& \textbf{Acc}$\uparrow$ & \textbf{Retain}$\uparrow$
& \textbf{Retain}$\uparrow$ \\
\midrule
\midrule
\textbf{-} & \textbf{LLaVA-Video 7B}
& 82.9 & 100.0\% 
& 62.8 & 100.0\% 
& 71.3 & 100.0\% 
& 60.4 & 100.0\% 
& 61.2 & 100.0\% 
& {-} \\
\cmidrule(lr){1-13}

\multirow{6}{*}{\textbf{70\%}}
& FastV~\cite{fastv}
& 81.3 & 98.1\% 
& 60.0 & 95.5\% 
& 64.2 & 90.0\% 
& 56.8 & 94.0\% 
& 57.5 & 94.0\% 
& 94.5\% \\
& VisionZip~\cite{yang2025visionzip}
& \underline{82.3} & \underline{99.3\%} 
& 61.3 & 97.6\% 
& \textbf{69.8} & \textbf{97.9\%} 
& 58.8 & 97.4\% 
& 57.7 & 94.3\% 
& \underline{97.4\%} \\
& G-Prune~\cite{jiang2025kind}
& 82.2 & 99.2\% 
& 61.7 & 98.2\% 
& 67.2 & 94.2\% 
& 57.8 & 95.7\% 
& 58.6 & 95.8\% 
& 96.7\% \\
\cmidrule(lr){2-13}
& STTM~\cite{sttm}
& 81.2 & 98.0\% 
& \underline{62.0} & \underline{98.7\%} 
& 68.8 & 96.5\% 
& 57.7 & 95.5\% 
& {-} & {-} 
& 97.2\% \\
& FrameFusion~\cite{fu2024framefusion}
& 81.1 & 97.8\% 
& 60.8 & 96.8\% 
& 67.3 & 94.4\% 
& \underline{59.3} & \underline{98.2\%} 
& \textbf{60.3} & \textbf{98.5\%} 
& 97.1\% \\
& \textbf{ForestPrune}
& \textbf{82.8} & \textbf{99.9\%} 
& \textbf{62.3} & \textbf{99.2\%} 
& \underline{69.3} & \underline{97.2\%} 
& \textbf{59.5} & \textbf{98.5\%} 
& \underline{59.8} & \underline{97.7\%} 
& \textbf{98.6\%} \\
\cmidrule(lr){1-13}

\multirow{6}{*}{\textbf{80\%}}
& FastV~\cite{fastv}
& 80.9 & 97.6\% 
& 57.7 & 91.9\% 
& 66.8 & 93.7\% 
& 55.6 & 92.1\% 
& 58.3 & 95.3\% 
& 94.3\% \\
& VisionZip~\cite{yang2025visionzip}
& \underline{81.9} & \underline{98.8\%} 
& \underline{60.0} & \underline{95.5\%} 
& \underline{67.8} & \underline{95.1\%} 
& \underline{57.5} & \underline{95.2\%} 
& \textbf{59.9} & \textbf{97.9\%} 
& \underline{96.6\%} \\
& G-Prune~\cite{jiang2025kind}
& 81.3 & 98.1\% 
& \underline{60.0} & \underline{95.5\%} 
& 65.5 & 91.9\% 
& 56.5 & 93.5\% 
& \underline{59.8} & \underline{97.7\%} 
& 95.4\% \\
\cmidrule(lr){2-13}
& STTM~\cite{sttm}
& 80.9 & 97.6\% 
& 59.8 & 95.2\% 
& 67.1 & 94.1\% 
& 55.9 & 92.6\% 
& {-} & {-} 
& 95.1\% \\
& FrameFusion~\cite{fu2024framefusion}
& 81.5 & 98.3\% 
& 59.9 & 95.4\% 
& 66.5 & 93.3\% 
& 56.9 & 94.2\% 
& 59.6 & 97.4\% 
& 95.8\% \\
& \textbf{ForestPrune}
& \textbf{82.4} & \textbf{99.4}\% 
& \textbf{60.5} & \textbf{96.3\%} 
& \textbf{68.5} & \textbf{96.1\%} 
& \textbf{57.6} & \textbf{95.4\%} 
& \textbf{59.9} & \textbf{97.9\%} 
& \textbf{97.1\%} \\
\cmidrule(lr){1-13}

\multirow{6}{*}{\textbf{90\%}}
& FastV~\cite{fastv}
& 73.2 & 88.3\% 
& 53.4 & 85.0\% 
& 58.5 & 82.1\% 
& 52.6 & 87.1\% 
& 51.3 & 83.8\% 
& 85.4\% \\
& VisionZip~\cite{yang2025visionzip}
& \underline{79.6} & \underline{96.0\%} 
& 57.0 & 90.8\% 
& 64.5 & 90.5\% 
& 51.9 & 85.9\% 
& 51.7 & 84.5\% 
& 90.0\% \\
& G-Prune~\cite{jiang2025kind}
& 78.4 & 94.6\% 
& \underline{57.7} & \underline{91.9\%} 
& 59.9 & 84.0\% 
& 52.1 & 86.3\% 
& 52.5 & 85.8\% 
& 88.8\% \\
\cmidrule(lr){2-13}
& STTM~\cite{sttm}
& 65.0 & 78.4\% 
& 56.0 & 89.2\% 
& \underline{64.8} & \underline{90.9\%} 
& 54.5 & 90.2\% 
& {-} & {-} 
& 86.6\% \\
& FrameFusion~\cite{fu2024framefusion}
& 78.9 & 95.2\% 
& 55.4 & 88.2\% 
& 60.3 & 84.6\% 
& \underline{54.7} & \underline{90.6\%} 
& \underline{57.2} & \underline{93.5\%} 
& \underline{90.5\%} \\
& \textbf{ForestPrune}
& \textbf{81.0} & \textbf{97.7\%} 
& \textbf{59.1} & \textbf{94.1\%} 
& \textbf{66.4} & \textbf{93.1\%} 
& \textbf{55.9} & \textbf{92.6\%} 
& \textbf{57.8} & \textbf{94.4\%} 
& \textbf{94.6\%} \\
\midrule
\midrule
\textbf{-} & \textbf{LLaVA-OV 7B}
& 80.1 & 100.0\% 
& 58.7 & 100.0\% 
& 64.4 & 100.0\% 
& 56.8 & 100.0\% 
& 58.2 & 100.0\% 
& {-} \\
\cmidrule(lr){1-13}

\multirow{6}{*}{\textbf{70\%}}

& FastV~\cite{fastv}
& 78.8 & 98.4\% 
& 55.7 & 94.9\% 
& 61.9 & 96.1\% 
& 55.5 & 97.7\% 
& 55.5 & 95.4\% 
& 96.6\% \\

& VisionZip~\cite{yang2025visionzip}
& 79.1 & 98.8\%
& 57.2 & 97.4\% 
& \underline{64.2} & \underline{99.7\%} 
& 55.9 & 98.4\%
& 56.2 & 96.6\% 
& 98.2\% \\

& G-Prune~\cite{jiang2025kind}
& 79.3 & 99.0\% 
& 57.4 & 97.8\%
& 63.2 & 98.1\% 
& \textbf{56.8} & \textbf{100.0\%} 
& 56.1 & 96.4\% 
&98.3\% \\

\cmidrule(lr){2-13}

& STTM~\cite{sttm}
& \underline{79.9} & \underline{99.8\%} 
& \textbf{58.7} & \textbf{100\%} 
& 63.7 & 98.9\% 
& 54.6 & 96.1\% 
& {-} & {-} 
& \underline{98.8\%} \\

& FrameFusion~\cite{fu2024framefusion}
& \textbf{80.3} & \textbf{100.2\%} 
& 57.2 & 97.4\% 
& 63.9 & 99.2\% 
& 55.1 & 97.0\% 
& \textbf{57.6} & \textbf{99.0\%} 
& 98.7\% \\

& \textbf{ForestPrune}
& 79.8 & 99.6\%
& \underline{57.7} & \underline{98.3\%} 
& \textbf{64.4} & \textbf{100.0\%} 
& \underline{56.3} & \underline{99.1\%} 
& \underline{57.3} & \underline{98.5\%} 
& \textbf{99.2\%} \\

\cmidrule(lr){1-13}

\multirow{6}{*}{\textbf{80\%}}

& FastV~\cite{fastv}
& 77.6 & 96.9\% 
& 52.9 & 90.1\% 
& 59.5 & 92.4\% 
& 52.7 & 92.8\% 
& 53.7 & 92.3\% 
& 93.1\% \\

& VisionZip~\cite{yang2025visionzip}
& 78.6 & 98.1\% 
& \underline{56.3} & \underline{95.9\%} 
&62.5 & 97.1\%
& 53.7 & 94.5\% 
& 56.2 & 96.6\% 
& 96.6\% \\

& G-Prune~\cite{jiang2025kind}
& 78.4 & 97.9\% 
& 55.8 & 95.1\% 
& 61.7 & 95.8\% 
& 52.5 & 92.4\% 
& 55.5 & 95.4\% 
& 95.5\% \\

\cmidrule(lr){2-13}

& STTM~\cite{sttm}
& 78.8 & 98.4\% 
& 56.1 & 95.6\% 
& \underline{63.7} & \underline{98.9\%} 
& 52.3 & 92.1\% 
& {-} & {-} 
& 96.5\% \\

& FrameFusion~\cite{fu2024framefusion}
& \textbf{79.6} & \textbf{99.4\%} 
& \underline{56.3} & \underline{95.9\%} 
& 61.7 & 95.8\% 
& \underline{54.4} & \underline{95.8\%} 
& \underline{57.4} & \underline{98.6\%} 
& \underline{97.2\%} \\

& \textbf{ForestPrune}
& \underline{79.2} & \underline{98.9\%} 
& \textbf{56.8} & \textbf{96.8\%} 
& \textbf{64.1} & \textbf{99.5\%} 
& \textbf{56.2} & \textbf{98.9\%} 
& \textbf{57.6} & \textbf{99.0\%} 
& \textbf{98.6\%} \\

\cmidrule(lr){1-13}

\multirow{6}{*}{\textbf{90\%}}

& FastV~\cite{fastv}
& 76.4 & 95.4\% 
& 50.9 & 86.7\% 
& 57.4 & 89.1\% 
& \underline{51.2} & \underline{90.1\%} 
& 53.0 & 91.1\% 
& 90.8\% \\

& VisionZip~\cite{yang2025visionzip}
& 74.4 & 92.9\% 
& 50.8 & 86.5\% 
& 58.7 & 91.2\% 
& 47.7 & 84.0\% 
& 52.0 & 89.4\% 
& 89.1\% \\

& G-Prune~\cite{jiang2025kind}
& 73.1 & 91.8\% 
& \underline{53.1} & \underline{90.5\%} 
& 53.1 & 82.5\% 
& 50.1 & 88.2\% 
& 52.3 & 89.9\% 
& 88.5\% \\

\cmidrule(lr){2-13}

& STTM~\cite{sttm}
& 72.4 & 90.4\% 
& 49.8 & 84.8\% 
& \underline{60.2} & \underline{93.5\%} 
& 50.1 & 88.2\% 
& {-} & {-} 
& 89.4\% \\

& FrameFusion~\cite{fu2024framefusion}
& \textbf{77.9} & \textbf{97.3\%} 
& 51.8 & 88.2\% 
& 57.5 & 89.3\% 
& 51.1 & 90.0\% 
& \underline{55.5} & \underline{95.4\%} 
& \underline{92.3\%} \\

& \textbf{ForestPrune}
& \underline{77.2} & \underline{96.4\%} 
& \textbf{55.8} & \textbf{95.1\%} 
& \textbf{61.3} & \textbf{95.2\%} 
& \textbf{55} & \textbf{96.8\%} 
& \textbf{55.6} & \textbf{95.5\%} 
& \textbf{95.8\%} \\

\bottomrule
\end{tabularx}}
\vspace{-0.2cm}
\end{table*}

\begin{table}[t]
\centering
\footnotesize
\setlength{\abovecaptionskip}{0.1cm}
\setlength{\belowcaptionskip}{0.2cm}
\caption{Comparison between LLaVA-Video using ForestPrune and existing SOTA MLLMs. 
Keeping the same amount of input tokens, we use ForestPrune to scale up the input video frames, and achieve obvious performance gains.}
\label{tab:bottleneck}
\resizebox{\linewidth}{!}{
\begin{tabularx}{\columnwidth}{@{} X c c c *{4}{c} @{}}  
\toprule
\multirow[c]{2}{*}{Model} &
\multirow[c]{2}{*}{Size} &
\multirow[c]{2}{*}{Frames} &
\multirow[c]{2}{*}{Input tokens} &
\multicolumn{1}{c}{VideoMME} &
\multicolumn{1}{c}{MLVU} \\
\cmidrule(lr){5-5}\cmidrule(lr){6-6}
& & & & Acc$\uparrow$ & Acc$\uparrow$ \\
\midrule
GPT-4V/4T ~\cite{gpt4}              & -  & 10 & - & 59.9 & 49.2 \\
Qwen2.5-VL ~\cite{Qwen2.5-vl}       & 7B & 64 & $>12544$ & 65.1 & 69.6 \\
Qwen3-VL ~\cite{qwen3technicalreport} & 8B & 2fps & $>12544$ & \textbf{71.4} & \textbf{78.1} \\
GLM-4.1V ~\cite{GLM-V-2025-GLM4.5V} & 9B & 64 & 16384 & \underline{68.2} & 71.5 \\
InternVL3 ~\cite{zhu2025internvl3}  & 8B & 64 & 16384 & 66.3 & 71.4 \\
InternVL3.5 ~\cite{Internvl3.5}     & 8B & 64 & 16384 & 66.0 & 70.2 \\
\midrule
LLaVA-Video~\cite{zhang2024video}             & 7B & 64  & 10816 & 62.8 & 71.3 \\
+ $\text{ForestPrune}_{\textbf{50\%}}$        & 7B & 128 & 10816 & 63.7 & 70.5 \\
+ $\text{ForestPrune}_{\textbf{75\%}}$        & 7B & 256 & 10816 & 64.2 & \underline{72.5} \\
+ $\text{ForestPrune}_{\textbf{87.5\%}}$      & 7B & 512 & 10816 & 66.5 & 72.2 \\
\bottomrule
\end{tabularx}}
\end{table}


\begin{table}[t]
\centering
\footnotesize
\setlength{\abovecaptionskip}{0.1cm}
\setlength{\belowcaptionskip}{0.2cm}
\setlength{\tabcolsep}{3pt}
\caption{Ablation study of the construction manners and numbers of the potential tree nodes $\mathbf{F}_{nod}$ for ForestPrune on LLaVA-Video.}
\label{tab:method}
\begin{tabularx}{\linewidth}{@{} c c *{6}{Y} @{}}
\toprule
\multirow[c]{2}{*}{Node} & \multirow[c]{2}{*}{FPrune} &
\multicolumn{2}{c}{NExT\text{-}QA} &
\multicolumn{2}{c}{VideoMME} &
\multicolumn{2}{c}{MLVU} \\
\cmidrule(lr){3-4}\cmidrule(lr){5-6}\cmidrule(lr){7-8}
 &  & Acc$\uparrow$ & Retain$\uparrow$
 & Acc$\uparrow$ & Retain$\uparrow$
 & Acc$\uparrow$ & Retain$\uparrow$ \\
\midrule
G-Prune & $\times$                   & 74.7 & 91.0\% & 57.7 & 91.9\% & 59.9 & 84.0\% \\
VisionZip & $\times$                & 76.5 & 91.7\% & 57.0 & 90.8\% & 64.5 & 90.5\% \\
$\times$     & $\checkmark$         & 80.1 & 96.0\% & \underline{58.9} & \underline{93.8\%} & 63.9 & 89.6\% \\
G-Prune & $\checkmark$               & \textbf{81.0 }& \textbf{97.1\%} & \textbf{59.1} & \textbf{94.1\%} & \underline{66.4} & \underline{93.1\%} \\
VisionZip & $\checkmark$            & \underline{80.6} & \underline{96.6\%} & 58.4 & 93.0\% & \textbf{66.7} & \textbf{93.5\%} \\
Random & $\checkmark$               & 80.1 & 96.0\% & 58.2 & 92.7\% & 65.6 & 92.0\% \\
\midrule
\multicolumn{8}{c}{\textbf{The keep ratio of tree nodes for each frame}}\\
\midrule
\multicolumn{2}{c}{100\%} & 80.1 & 96.0\% & 58.9 & 93.8\% & 63.9 & 89.6\% \\
\multicolumn{2}{c}{70\%}  & 80.6 & 97.2\% & 58.9 & 93.8\% & 66.4 & 93.1\% \\
\multicolumn{2}{c}{60\%}  & 80.7 & 97.3\% & 58.9 & 93.8\% & 66.4 & 93.1\% \\
\multicolumn{2}{c}{50\%}  & 80.8 & 97.5\% & 58.8 & 93.6\% & 66.4 & 93.1\% \\
\bottomrule
\end{tabularx}
\end{table}

\vspace{-0.1cm}
\subsection{Implement Detail}
\vspace{-0.2cm}
We apply our ForestPrune to two representative video MLLMs, namely LLaVA-Video-7B~\cite{zhang2024video} and LLaVA-OneVision-7B~\cite{Llava-onevision}. 
Following their default settings, we input 64 frames with a $13\times13$ patch tokens on LLaVA-Video, and input 32 frames with a $14\times14$ tokens on LLaVA-OneVision.
We primarily set the semantic thresholds $\tau_s$ to 0.9 and spatial thresholds $\tau_p$ to 0.8, and their best settings are defined based on MLLM.
By default, we keep the number of nodes at 50\% of the original number.
And we compare ForestPrune with other image and video pruning methods. 
For image based method, we compare the pruning based methods of FastV~\cite{fastv} and G-Prune~\cite{jiang2025kind} and the hybrid method of VisionZip~\cite{yang2025visionzip} that adopts both token pruning and merging. 
For video based methods, we compare two token merging methods, namely STTM~\cite{sttm} and FrameFusion~\cite{fu2024framefusion}. 
Considering the differences in base MLLMs, experimental settings and the lack of results, we follow previous works~\cite{sttm, fu2024framefusion} to faithfully reproduce their performance in our experiments using their default code projects. 

\vspace{-0.1cm}
\subsection{Benchmarks and Metrics}
\vspace{-0.2cm}
To validate ForestPrune, we conduct extensive experiments on five public video understanding benchmarks: NExT-QA~\cite{NExT-QA}, MVBench~\cite{Mvbench}, VideoMME~\cite{Video-mme}, MLVU~\cite{MLVU}, and LongVideoBench~\cite{LongVideoBench}.
NExT-QA is a multiple-choice video QA benchmark with short clips averaging about 44 seconds. 
MVBench assesses multi-skill short-video understanding with clips averaging roughly 16 seconds. 
VideoMME is a comprehensive benchmark, spanning short, medium and long videos. 
MLVU targets at long videos ranging from 3 minutes to 2 hours, with an average around 15 minutes. 
LongVideoBench covers diverse video pieces with four groups, which are 8–15 seconds, 15–60 seconds, 3–10 minutes, and 15–60 minutes, respectively.

\vspace{-0.1cm}
\subsection{Quantitative Analysis}
\vspace{-0.2cm}
\textbf{Performance Comparison with SOTA methods.}
We first compare ForestPrune to existing visual compression methods on LLaVA-Video and LLaVA-OV in Tab \ref{tab:main_results}.
The performance of these base MLLMs without token compression are used as reference. 
From these results, we can first observe that image-centric methods can still performs well in these video tasks. 
For instance, VisionZip can outperform other video-based methods under the setting of LLaVA-Video-70\%. 
And G-Prune also performs well on LLaVA-OV, \emph{e.g.,} 98.3\% for 70\% pruning ratio.
However, under the extremely high pruning ratio, their retained performance still drop greatly, \emph{e.g.}, -9.1\% and -7.4\% by FastV and VisionZip from 70\% to 90\% ratios. 
Besides, it can be also seen that the video-oriented methods, such as STTM and FrameFusion, are also inferior in high-ratio settings, \emph{e.g.,} 90\%. 
In stark contrast, our ForestPrune not only shows better performance than the compared methods, but also retain high accuracies under the high-ratio compression cases, \emph{e.g.}, 94.6\% and 95.8\% for LLaVA-Video and LLaVA-OV under the 90\% setting. 
These results well confirm the effectiveness of ForestPrune towards high-ratio video compression for MLLMs.

In addition, we also use ForestPrune to scale up the input frames of LLaVA-Video and compare it with existing SOTA MLLMs on VideoMME and MLVU, as reported in Tab \ref{tab:bottleneck}.
From the table, we can first observe that ForestPrune can scale up the input frames while maintaining the same number of input tokens. 
On both the VideoMME and MLVU benchmarks, ForestPrune significantly improves LLaVA-Video's performance by processing more frames. 
More importantly, we help LLaVA-Video outperform the most advanced video MLLMs such as InternVL3 and InternVL3.5~\cite{zhu2025internvl3, Internvl3.5}.
Overall, ForestPrune not only maintains the model's performance well under high pruning rates, but also improves model performance by scaling up the input frames.

\begin{figure}[t]
    \centering
    \includegraphics[width=\linewidth]{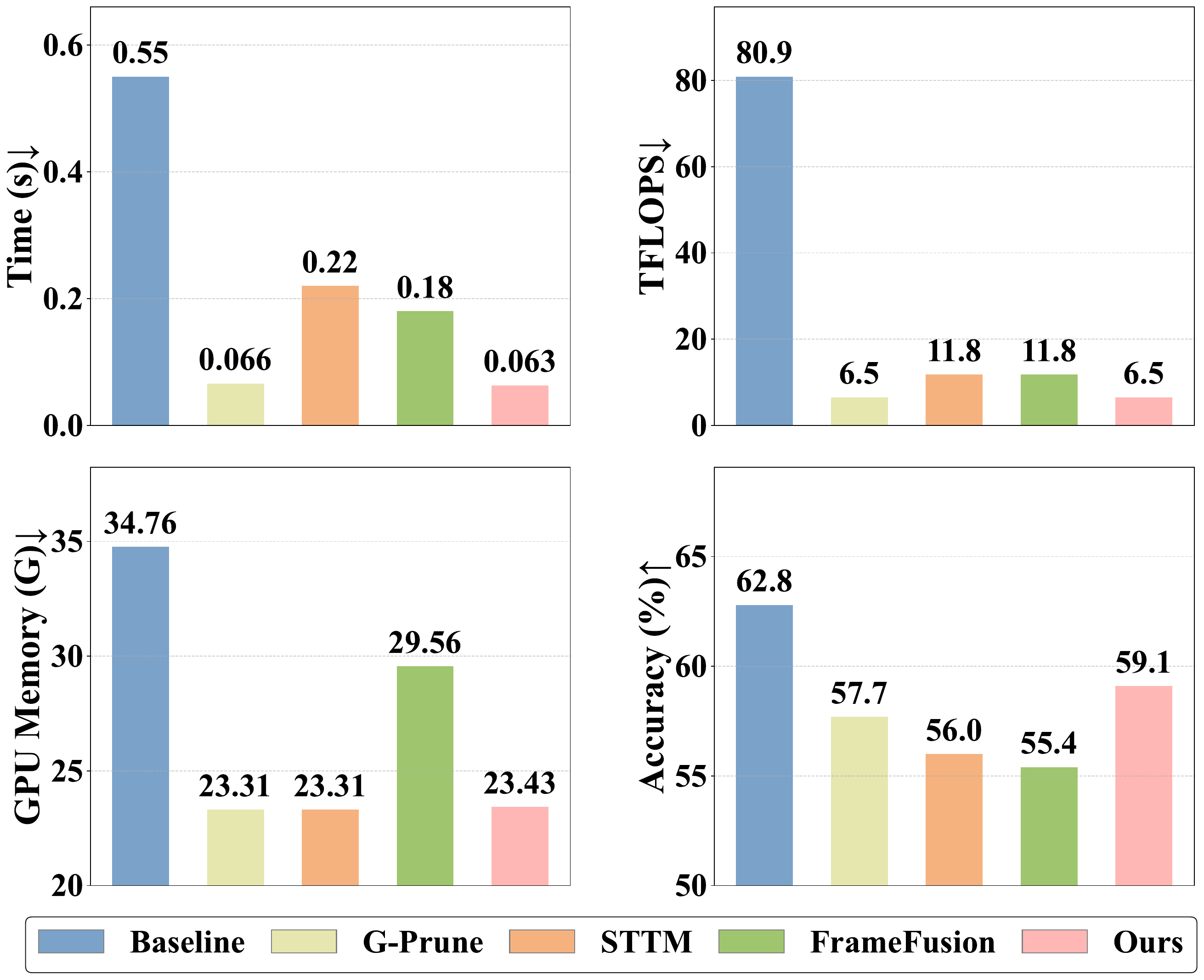}
    \vspace{-0.3cm}
    \setlength{\abovecaptionskip}{0.cm}
    \caption{Efficiency comparison between ForestPrune and three existing methods. It is using LLaVA-Video with 90\% compression ratio. \emph{GPU memory} records the peak usage.}
    \label{fig:effiency}
    \vspace{-0.2cm}
\end{figure}

\textbf{Efficiency Comparison with SOTA methods.}
We make efficiency comparisons between ForestPrune and a bunch of image and video pruning methods under the ratio of 90\%, as shown in Fig \ref{fig:effiency}. 
From these charts, we can first observe that all methods can well shorten the prefilling time of LLaVA-Video via reducing the amount of video tokens. 
In particular, the prefilling efficiency of the pre-compression methods, such as G-Prune and our ForestPrune, is more obvious, since they compress tokens before MLLM's encoding. 
Similarly, these methods are also superior in the reductions of computation complexity and GPU memory overhead. 
For the post-compression methods, \emph{i.e.}, compressing tokens in MLLMs due to their token evaluations in MLLMs, \emph{e.g.}, FrameFusion. 
In particular, our ForestPrune is not only one of the most efficient approaches, but also, as mentioned in the previous paragraph, keeps the highest accuracies. 
These results well confirm the efficiency of ForestPrune.

\textbf{Ablation Study.}
In Tab \ref{tab:method}, we first ablate the collection and number of candidate tree nodes, \emph{i.e.}, $\mathbf{F}_{nod}$ in Sec \ref{overview}. 
Here, the performance of G-Prune and VisionZip are used as reference. 
From the first block of the table, we can observe that the selection of tree nodes from each frame has a certain impact on ForestPrune. 
For instance, using all frame tokens will be more difficult to construct the spatial-temporal trees, resulting in sub-optimal performance, \emph{i.e.}, $\times$ in \emph{Image}. 
A better selection of tree nodes can help to improve performance, \emph{e.g.}, $\langle \text{G-Prune}, \checkmark\rangle$, but it has not great difference in methodologies, \emph{e.g.}, \emph{Random} is also feasible. 
Similarly, from the second block, we can also see that keeping a smaller number of frame tokens are beneficial in tree construction. 
But it does not much matters the results.


\begin{table}[t]
\centering
\footnotesize
\setlength{\abovecaptionskip}{0.1cm}
\setlength{\belowcaptionskip}{0.2cm}
\setlength{\tabcolsep}{3pt}
\caption{Ablation of the semantic and spatial hyper-parameters of ForestPrune on LLaVA-Video-7B, \emph{i.e.}, $\tau_s$ and $\tau_p$.}
\label{tab:parameter}
\begin{tabularx}{\linewidth}{@{} c c *{6}{Y} @{}}
\toprule
\multirow[c]{2}{*}{$\tau_s$} &
\multirow[c]{2}{*}{$\tau_p$} &
\multicolumn{2}{c}{NExT\text{-}QA} &
\multicolumn{2}{c}{VideoMME} &
\multicolumn{2}{c}{MLVU} \\
\cmidrule(lr){3-4}\cmidrule(lr){5-6}\cmidrule(lr){7-8}
 & & Acc$\uparrow$ & Retain$\uparrow$
   & Acc$\uparrow$ & Retain$\uparrow$
   & Acc$\uparrow$ & Retain$\uparrow$ \\
\midrule
0.95 & 0.80 & 80.7 & 97.3\% & 58.8 & 93.6\% & 66.4 & 93.1\% \\
0.90 & 0.80 & 80.9 & 97.6\% & 58.8 & 93.6\% & 66.4 & 93.1\% \\
0.80 & 0.80 & 79.8 & 96.3\% & 58.8 & 93.6\% & 66.4 & 93.1\%  \\
\midrule
0.80 & 0.50 & 79.4 & 95.8\% & 58.5 & 93.2\% & 65.8 & 92.3\% \\
0.80 & 0.60 & 79.5 & 95.9\% & 58.5 & 93.2\% & 65.3 & 91.6\% \\
0.80 & 0.90 & 80.2 & 96.7\% & 58.9 & 93.8\% & 66.4 & 93.1\% \\
0.80 & 0.95 & 81.0 & 97.7\% & 58.8 & 93.6\% & 66.1 & 92.7\% \\
\bottomrule
\end{tabularx}
\vspace{+0.1cm}
\end{table}

In Tab \ref{tab:parameter}, we ablate the two hyper-parameters $\tau_s, \tau_p$ of ForestPrune, \emph{i.e.}, the semantic and spatial constraints, respectively. 
These two hyper-parameters also affect the qualities of constructed trees. 
We can see that via using a large threshold values, ForestPrune can construct better token trees with fewer noisy connections, thereby facilitating the final token pruning. 
We can also see that using a smaller spatial threshold increases noisy connections in the token tree, leading to slight drops in performance. 
Similar to Tab \ref{tab:method}, when the thresholds reach a certain value, ForestPrune is not sensitive to their slight changes, \emph{e.g.}, the performance of 60\% and 50\% are almost the same. 
Overall, these results well confirm the effectiveness and robustness of our ForestPrune towards its designs and settings.

\begin{figure*}[t]
\label{fig:forest}
    \centering
    \includegraphics[width=\textwidth]{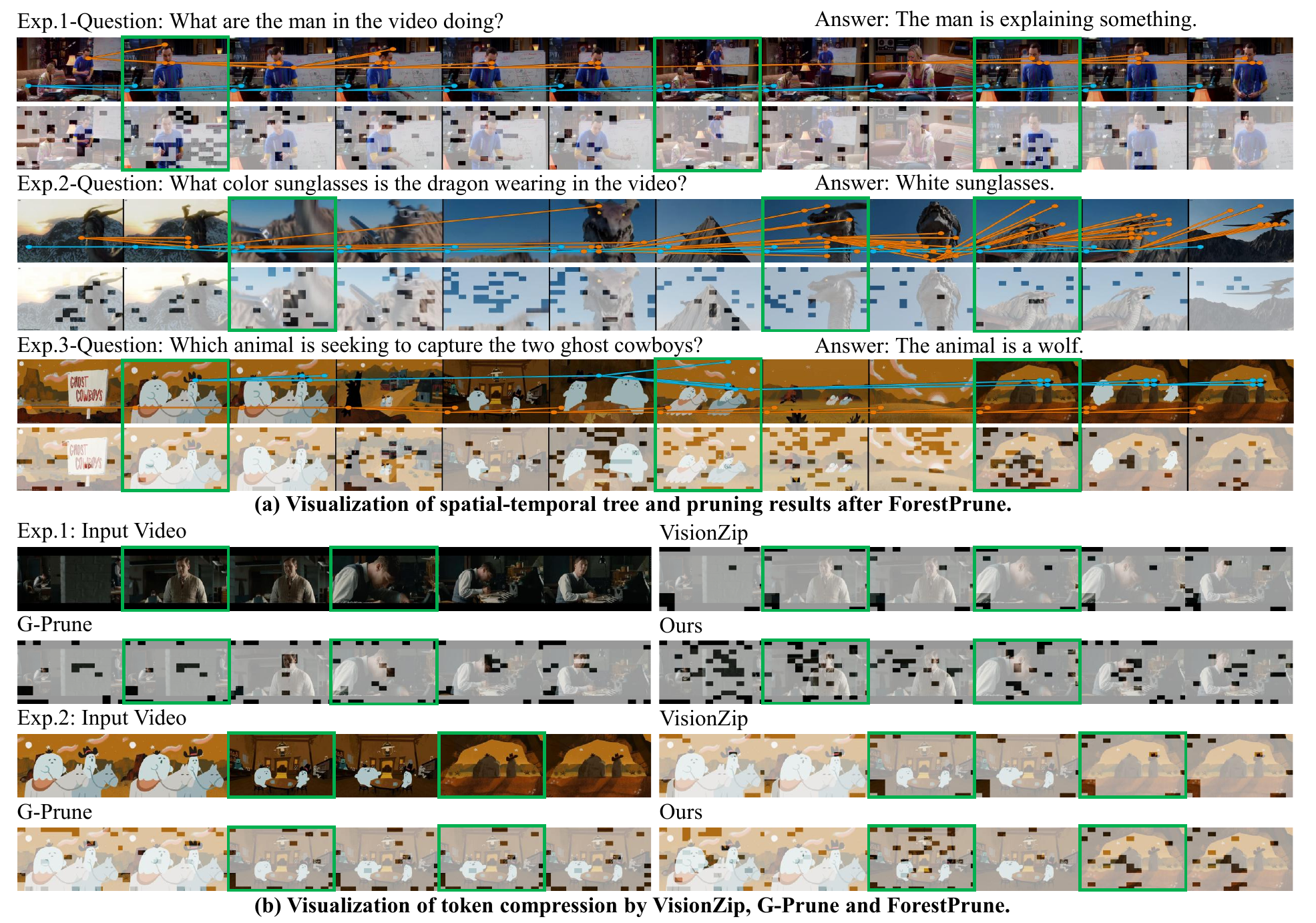}
    \vspace{-0.5cm}
    \setlength{\abovecaptionskip}{0.2cm}
    \caption{Visualized results of ForestPrune. 
    Subfigure-(a) shows the spatial-temporal tree built by ForestPrune and the compression results, which show ForestPrune's global spatial-temporal modeling capabilities. 
    Subfigure-(b) shows the compression results of ForestPrune, G-Prune, and VisionZip, showcasing ForestPrune's ability to reduce temporal redundancy compared to image compression methods. 
    The \textcolor{orange}{ORANGE} and \textcolor{blue}{BLUE} trees are the spatial-temporal trees. 
    Frames with scene changes occur are shown with \textcolor{green}{GREEN} boxes.}
    \label{fig:qualitative}
    \vspace{-0.5cm}
\end{figure*}

\vspace{-0.1cm}
\subsection{Qualitative Analysis}
\vspace{-0.2cm}
To gain deeper insight into ForestPrune's process of building a spatial-temporal forest, we visualize its forest construction and pruning results, as shown in Fig \ref{fig:qualitative}(a). 
 As observed, the spatial-temporal tree can span multiple frames, allowing each tree to represent global temporal information. 
And we can see that the nodes in a same tree are similar, \emph{e.g.}, the orange tree in the Exp.1 which capture human's face. 
This proves that the spatial-temporal trees can capture similar semantic information across video. 
In addition, it can be seen that due to the spatial constraint of the spatial-temporal trees, the trees can only be constructed within a limited area, \emph{e.g.}, the orange and blue trees in the Exp.3. 
This reduces noisy in the tree, making the spatial-temporal tree more accurate.

In the Fig \ref{fig:qualitative}(b), we visualize the compression results of ForestPrune, G-Prune, and VisionZip on different video examples. 
First, ForestPrune gradually reduces the number of tokens per frame when consecutive scenes are similar. 
Afterward, when the scene changes, the number of tokens retained by ForestPrune increases rapidly. 
In addition, if a scene reappear, the number of tokens retained by ForestPrune increased slightly. 
These phenomena indicate that ForestPrune effectively preserves diversity information in the video. 
In contrast, G-Prune and VisionZip exhibit substantial temporal redundancy in the retained tokens. 
For instance, in the Exp.1 and Exp.2, ForestPrune retains a more diverse range of tokens, while G-Prune and VisionZip repeatedly retain nearly the same tokens across similar frames. Overall, these results well confirm the merits of ForestPrune in spatial-temporal modeling.
\vspace{-0.1cm}
\section{Conclusion}
\label{sec:conclusion}
\vspace{-0.2cm}
In this paper, we introduce ForestPrune, a novel and training-free high-ratio compression method for video MLLMs. 
ForestPrune constructs token forests across video frames based on the semantic, spatial and temporal constraints. 
Extensive comparisons with existing compression methods across various benchmarks demonstrate that ForestPrune well retain the video MLLMs performance while reducing a large number of redundant tokens, fully validating its designs and our motivations.

\vspace{-0.1cm}
\section{Acknowledgments}
\vspace{-0.2cm} 
This work was supported by the National Key Research and Development Program of China (No.2025YFE0113500), the National Science Fund for Distinguished Young Scholars (No.62525605), the National Natural Science Foundation of China (No.U25B2066, No.U22B2051, No.62572407) , Fujian Province Special Science and Technology Program (No.2025H0041).


{
    \small
    \bibliographystyle{ieeenat_fullname}
    \bibliography{main}
}


\end{document}